\documentclass[a4paper]{article}

\usepackage{arxiv}
\usepackage[T1]{fontenc}
\usepackage[utf8]{inputenc}%

\usepackage{graphicx}
\usepackage{hyperref}

\usepackage{xcolor}

\usepackage{authblk}
\usepackage{geometry}
\usepackage{setspace}
\usepackage{verbatim}
\usepackage[utf8]{inputenc}
\usepackage{amsmath}
\usepackage{amssymb}
\usepackage{gensymb}
\usepackage{verbatim} % env for block comment 
\usepackage{ragged2e} % para usar flusleft
\usepackage[frozencache,cachedir=.]{minted}
\usepackage{amsfonts}
\usepackage{cancel}

\usepackage[sorting=none]{biblatex}
\usepackage{listings}

\usepackage[utf8]{inputenc}

\usepackage{tabularx} %
\usepackage{multirow}
\usepackage{makecell}
\newcommand{\abstractinenglishname}{Abstract}

\renewenvironment{abstract}{%
        \begin{center}
	\begin{minipage}{14cm}
	{\textbf{\abstractname:}}
}{
        \end{minipage}
	\end{center}
}
\newenvironment{abstractinenglish}{
        \def\abstractname{\abstractinenglishname}
	\begin{abstract}
}{
        \end{abstract}
}

\title{Leveraging Ontologies to Document Bias in Data}

\author{Mayra Russo \thanks{\textcolor{red}{A version of this work has been accepted and will be presented at AEQUITAS 2024: Workshop on Fairness and Bias in AI | co-located with ECAI 2023, Santiago de Compostela, Spain. This contribution will be published in the CEUR Workshop Proceedings.}} \\
\thanks{email: mrusso@l3s.de,} $^1$}
\affil{$^1$Leibniz Universität Hannover, Germany}
\date{}

\author{Maria-Esther Vidal \thanks{email:maria.vidal@tib.eu} $^1$,$^2$}
\affil{$^2$ TIB Leibniz Information Center for Science and Technology, Hannover, Germany}

\usepackage{fancyhdr}
\begin{document}

\maketitle

\begin{abstractinenglish}
\emph{Machine Learning (ML) systems are capable of reproducing and often amplifying undesired biases. This puts emphasis on the importance of operating under practices that enable the study and understanding of the intrinsic characteristics of ML pipelines, prompting the emergence of documentation frameworks with the idea that ``any remedy for bias starts with awareness of its existence''. However, a resource that can formally describe these pipelines in terms of biases detected is still amiss. To fill this gap, we present the Doc-BiasO ontology, a resource that aims to create an integrated vocabulary of biases defined in the \textit{fair-ML} literature and their measures, as well as to incorporate relevant terminology and the relationships between them. Overseeing ontology engineering best practices, we re-use existing vocabulary on machine learning and AI, to foster knowledge sharing and interoperability between the actors concerned with its research, development, regulation, among others. Overall, our main objective is to contribute towards clarifying existing terminology on bias research as it rapidly expands to all areas of AI and to improve the interpretation of bias in data and downstream impact.}
\end{abstractinenglish}

\section{Introduction}
\label{sec:intro}
\begin{figure}[!t]
 \vspace{-0.5cm}
    \centering    {\includegraphics[width=10cm,height=8cm,keepaspectratio]{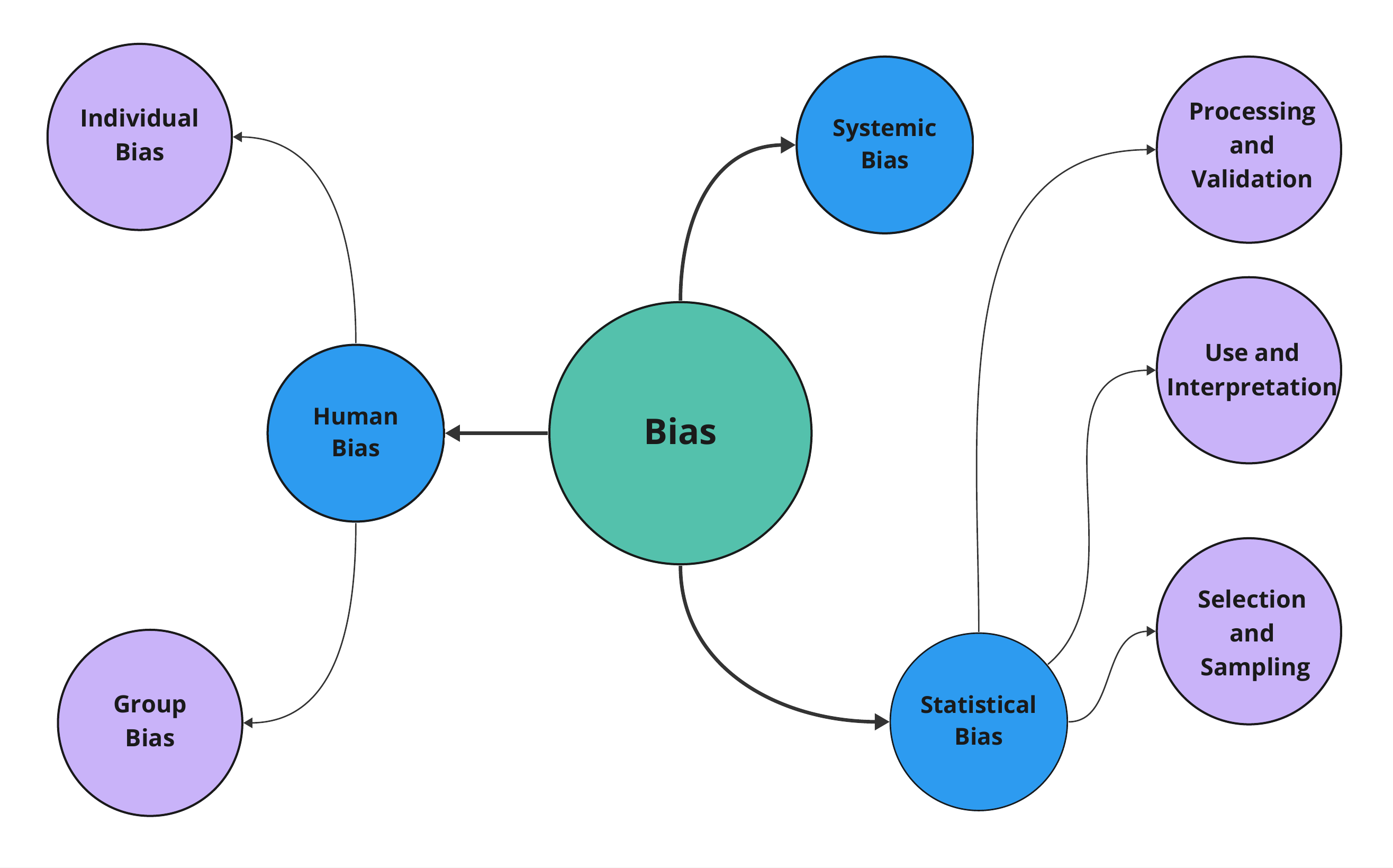}}    
\caption{\textbf{Types of Bias.} Core categories of bias in relation to AI systems as per the NIST report \cite{933006}.}
\label{fig:nist}
     \vspace{-0.5cm}
\end{figure}

The breakthroughs and benefits attributed to big data and, consequently, to machine learning (ML) - or \textit{AI}- systems \cite{10.1145/3465416.3483305,10.1145/3458742}, have also resulted in making prevalent how these systems are capable of producing unexpected, biased, and in some cases, undesirable output \cite{BaezaYates2013BigDO,barocas-hardt-narayanan,Barocas2016BigDD}. Seminal work on bias (i.e., prejudice for, or against one person, or group, especially in a way considered to be unfair) in the context of ML systems demonstrates how facial recognition tools and popular search engines can exacerbate demographic disparities, worsening the marginalization of minorities at the individual and group level \cite{Buolamwini2018GenderSI,Noble2018AlgorithmsOO}. Further, biases in news recommenders and social media feeds actively play a role in conditioning and manipulating people's behavior and amplifying individual and public opinion polarization \cite{BaezaYates2020BiasOT,BaezaYates2020BiasesOS}. In this context, the last few years have seen the consolidation of the \textit{Trustworthy AI} framework, led in large part by regulatory bodies \cite{doi:10.1080/19460171.2024.2315431}, with the objective of guiding commercial AI development to proactively account for ethical, legal, and technical dimensions \cite{hleg}. Furthermore, this framework is also accompanied by the call to establish standards across the field in order to ensure AI systems are safe, secure and fair upon deployment \cite{hleg}. In terms of AI bias, many efforts have been concentrated in devising methods that can improve its identification, understanding, measurement, and mitigation \cite{Alvarez2024PolicyAA}. For example, the special publication prepared by the National Institute of Standards and Technology (NIST) proposes a thorough, however not exhaustive, categorization of different types of bias in AI beyond common computational definitions (see Figure \ref{fig:nist} for core hierarchy) \cite{933006}. In this same direction, some scholars advocate for practices that account for the characteristics of ML pipelines (i.e., datasets,
ML algorithms, and user interaction loop) \cite{Mehrabi2019ASO} to enable actors concerned with its research, development, regulation, and use, to inspect all the actions performed across the engineering process, with the objective to increase trust placed not only on the development processes, but on the systems themselves \cite{10.1145/3458723,Raji2019ABOUTMA,10.1145/3488717,10.1145/3351095.3372873}.\\
In addition to human-readable (i.e., textual descriptions in a format that humans can read and understand) documentation frameworks for machine learning pipelines \cite{bender-friedman-2018-data,10.1145/3458723,10.1145/3287560.3287596}, Semantic Web technologies (e.g., ontologies, knowledge graphs) can also play a crucial role in enhancing the accuracy and interpretability of ML systems \cite{FERNANDEZ2023100808}, as well as to perform ``bias assessment, representation, and mitigation'' tasks \cite{ReyeroLobo2022SemanticWT}, in a way that is also machine-readable (i.e., makes available a fine-grained description of data in a format manageable by computers). This characteristic improves the findability,  accessibility, interoperability and reusability (\textit{FAIR}) of data-centric resources in the Web \cite{10.1145/3528574,wilkinson2016fair}. Ontologies to model existing ML fairness metrics \cite{10.1145/3514094.3534137,Franklin2023AnOF}, as well as the semantic specifications to catalog risks in terms of compliance and conformance of AI systems under the EU's AI Act\footnote{Annex III, European Council position} \cite{i-semantics/GolpayeganiP022,10.1145/3593013.3594050} have been proposed, however, a resource that can formally describe ML pipelines, and provides a vocabulary to characterize them in terms of measured biases is still amiss. 

\textbf{Proposed Solution} We propose an ontology-driven approach to describe and document biases detected across machine learning pipelines. Here, we refer to documentation as the process of generating metadata represented in formats understandable by humans and also by machines \cite{cacm/NoyG23}, where formal data models like ontologies and controlled vocabularies provide standardized concepts for expressing this metadata. Our ontology, \textit{Doc-BiasO}, is a resource developed with the objective to introduce an integrated vocabulary system of ML-related biases as defined in the literature and their measures; represent their relationships with other relevant terminology, i.e., datasets, ML systems, fairness, harm, risk; and semantically annotate ML pipelines based on bias measures values.
The version presented here has 389 classes, 72 object properties, and 28 data properties. 

\textbf{Contributions:} Concisely, our contributions are the following:
\begin{enumerate}
\item Doc-BiasO, an integrated vocabulary system of ML-related biases; 
    \item an ontology-based approach to document bias in ML pipelines; 
    \item a technical evaluation of Doc-BiasO.
\end{enumerate}

The remainder of this paper is structured as follows: Section \ref{sec:related} introduces relevant Semantic Web concepts and presents a review of related literature. Section \ref{sec:design}, details the design of Doc-BiasO. The results of the evaluation are reported in Section \ref{sec:eval}. Finally, Section \ref{sec:conclusion} outlines our conclusions and future lines of work.

\section{Background}
\label{sec:related}
\paragraph{Ontologies and Machine Learning}

Gruber \cite{gruber_toward_1995} defines an ontology as a formal, explicit specification of a shared conceptualization that is characterized by high semantic expressiveness required for increased complexity. Ontologies include abstract concepts or classes, represented as nodes, and predicates representing the relations of these classes, edges in an ontology; the meaning of the predicates is represented using rules. Ontologies are specified using knowledge representation models, making the expressiveness of the ontology dependent on the expressive power of the representation model. The Resource Description Framework (RDF)\footnote{\url{https://www.w3.org/RDF/}} enables the description of entities in terms of classes and properties; while subsumption relations between classes and properties can be modelled with the RDF Schema (RDFS).\footnote{\url{https://www.w3.org/TR/rdf12-schema/}} More expressive formalisms like the Ontology Web Language (OWL)\footnote{\url{https://www.w3.org/OWL/}} make available a larger number of operators which enable the representation not only of classes, properties, and subsumption relations, but also class and property constraints, general equivalence relations, and restrictions of cardinality. Several examples of the usefulness of context-aware ontologies for bias awareness and mitigation in ML systems are explored in the work presented in \cite{ReyeroLobo2022SemanticWT}. 

In the context of bias modelling, the \textbf{Bias Ontology Design Pattern (BODP)} \cite{Kaushik2021} is one of the first works to propose a formalization for the bias concept. Its objective is to capture a high-level representation of bias as an abstract term and not necessarily in the context of ML systems. We re-use part of BODP as building block, however Doc-BiasO has a different scope and intended use. Similar to our work, the \textbf{fairness metrics ontology (FMO)} \cite{10.1145/3514094.3534137,Franklin2023AnOF} models fairness metrics (\texttt{fmo:fairness\_metric}) from the literature and relates them to their use-case. The conceptualization of bias and fairness in relation to ML systems are often intertwined; however, distinctions between both concepts need to be made explicit, as they are not always used in conjunction, nor to study the same phenomena \cite{10.1145/3404193,10.1145/3588433}. Fairness, in relation to ML (\textit{fair-ML}), takes the form of algorithmic interventions that incorporate mathematical formalizations of moral or legal notions for the fair treatment of different populations into ML pipelines. These interventions aim to prompt ML models to satisfy statistical non-discrimination criterion for a given subpopulation \cite{barocas-hardt-narayanan}. Specifically, we propose a descriptive vocabulary that can be used and incorporated into varying frameworks as needed and that can be extended to further semi-automatize documentation tasks. Moreover, our focus is on modelling biases in data identified in the literature and the existing measures defined to detect it. These are concepts and relations that are not made explicit in the current version of FMO. As we consider both ontologies to be complementary, we re-use FMO to foster the development of a comprehensive vocabulary that provides coverage of terminology that pertains to the responsible development of ML systems. We follow the same approach with the \textbf{AI Risk Ontology (AIRO)} \cite{i-semantics/GolpayeganiP022}, and by-effect, the \textbf{Vocabulary of AI Risks (VAIR)} \cite{10.1145/3593013.3594050}; in this case, risk in relation to ML systems, under the broader label of AI, is defined as systems that are likely to cause serious harms to health, safety, or fundamental rights of individuals as per European Union (EU) Law. These works are ontology-driven approaches to account for the compliance and conformance of AI systems under the EU's AI Act's specifications\footnote{Annex III, European Council position}. Specifically, AIRO is a modular ontology created to identity whether an AI system is classified as high-risk, whilst VAIR provides semantic specifications for cataloging AI risks, re-using core concepts in AIRO (e.g., \texttt{airo\#Risk}, \texttt{airo\#Consequence}). Lastly, \cite{10.1145/3593013.3594059} proposes a descriptive framework \textbf{(ACROCPoLis)} to describe ML systems and their societal impact by making explicit the interrelations and diverging perspectives of relevant stakeholders (individuals, groups of people, institutions). While this is beyond the scope of our work, should the conceptual model be formalized and made publically available, a study for re-use and extension of Doc-BiasO ontology would be undertaken for a future iteration. 

The Semantic Web community has also proposed other technical solutions to improve the interpretability and transparency of machine learning pipelines. The provenance ontology PROV-O \cite{733f89c65e4844f9aabcae1c276a5602} enables the representation of provenance information generated by different entities, and can be easily applied to multiple contexts. Standard schemas for data mining and machine learning algorithms, such as the Machine Learning Schema (MLS) ontology \cite{MLSchema}, and the Description of a Model (DOAM)\footnote{\url{https://www.openriskmanual.org/ns/doam/index-en.html}} ontology, provide fine-grained vocabularies to represent ML models characteristics. Moreover, the issue of reproducibility in ML has also been addressed \cite{Albertoni2023ReproducibilityOM}. Correspondingly, the Data Catalog Vocabulary (DCAT) \cite{Albertoni2023TheWD} enables  the fine-grained description of datasets and data services in a catalog using a controlled and rich vocabulary. %By extension, the Data Quality Vocabulary (DQV) \cite{Albertoni2020IntroducingTD} provides a framework and vocabulary to assess the  quality of a dataset, offering an extensive catalog of quality metrics. 
Adhering to ontology engineering best practices \cite{gruber_toward_1995}, all these ontologies and vocabularies have been re-used in the composition of Doc-BiasO. 

\paragraph{Documentation Frameworks and Machine Learning}
The opaqueness of the inner processes of ML systems can hinder the understanding of how they work. \cite{bender-friedman-2018-data, 10.1145/3458723,10.1145/3287560.3287596,Hupont2024}, thus advocate for the production of value oriented, human-readable documentation for datasets (Data Statements for Natural Language Processing, Datasheets for Datasets), ML models (Model Cards for Model Reporting and Use Case Cards). Doc-BiasO aims to follow their stride by combining the different components of the ML pipeline (input, model, output data) to produce comprehensive descriptions in human- and machine-readable format of data-driven pipelines. Other documentation approaches, such as 
Sun et al. \cite{10.1145/3357384.3357853} introduce a tool to assess fitness for use of datasets. This automated data exploration tool delimits its focus to three dimensions: representativeness, bias, and correctness.  In a similar line, \cite{10.1007/s11263-022-01625-5} introduces a bias visualization tool for computer vision datasets. This exploration tool narrows down their assessment to three sets of metrics: object-based, gender-based and geography-based dimensions. Further, interactive tools-- developed by industries-- (e.g., \cite{google,hug,facebook}) enable dataset exploration, visualization, and comparison. The extensible and modular design of Doc-BiasO, allows users to describe and document their data-driven pipelines, and seamlessly incorporates additional descriptive dimensions and components as needed. Further, the underlying knowledge-driven framework prompts the integration and fine-grained description of multiple data sources, and leverages reasoning capabilities for enhanced data analytics. 

\begin{table}[!t]
\centering
\begin{tabularx}{.98\textwidth}{|l|X|}
\hline
\multicolumn{1}{|c|}{\textbf{Concept}} & \multicolumn{1}{c|}{\textbf{Definition}}   \\ \hline
Bias   & \begin{tabular}[c]{@{}l@{}} A concentration on, or interest in one particular area or subject.\\ Whilst a more value-laden definition, conceptualizes bias as prejudice\\for, or against one person, or group,  especially in a way considered\\to be unfair \cite{Pitoura2020}.  \end{tabular}  \\ \hline
Application  & \begin{tabular}[c]{@{}l@{}} The use, purpose or application of a machine learning system.\\ Examples include, recommenders, speech recognition, etc.\end{tabular}  \\ \hline
ML Task  & \begin{tabular}[c]{@{}l@{}} Task or ML Problem is the formal description of a process that needs\\to be completed (e.g., based on inputs and\\(outputs) \cite{MLSchema}. \end{tabular}  \\ \hline
Dataset  & \begin{tabular}[c]{@{}l@{}} A collection of data, published or curated by a single source, and\\available for access or download in one or more  \\ representations \cite{MLSchema}. \end{tabular}  \\ \hline
Harm  & \begin{tabular}[c]{@{}l@{}} Adverse lived experiences resulting from an ML system’s deployment\\and operation in the world \cite{10.1145/3600211.3604674,Shelby2022SociotechnicalHO}. \end{tabular}  \\ \hline
Bias Measure  & \begin{tabular}[c]{@{}l@{}} A quantitative metric or indicator that assesses the presence and\\extent of bias in a particular context, via predefined\\thresholds \cite{innovations/DworkHPRZ12}. \end{tabular}  \\ \hline
\end{tabularx}
\caption{\textbf{Core Concepts in Bias in Machine Learning.}}
\label{tab:concepts}
\end{table}

\section{Design and Implementation}
\label{sec:design}
In this section, we describe the design stages of Doc-BiasO. We also describe its implementation and include an example of an instance.

\subsection{Scoping out the Coverage of Doc-BiasO}
\label{sec:scope}
To determine the scope of our ontology, we perform a domain and content analysis following a hybrid strategy. On the one hand, through our own position within a research project on bias in relation to ML systems\footnote{\url{https://nobias-project.eu/}}, 
we have held fruitful discussions with experts researching different dimensions of bias from a multidisciplinary and critical point of view, i.e., \cite{10.1145/3625007.3627491,Alvarez2024PolicyAA}. Further, these discussions have helped identify what concepts make up our universe of discourse, for instance, bias, ML model, dataset, task, application, fairness, harms, risks; as well as how these concepts interact or relate to each other. In Table \ref{tab:concepts}, we have summarized the core concepts defined in our ontology. Each of these concepts represents the top-most abstract concept in a hierarchy of terms, with less abstract or more concrete concepts being defined as the ontology grows to give a broader coverage. For example, \texttt{Bias} is the most abstract representation, while \texttt{Representation Bias} is a more concrete type of bias. 

The exchanges with researchers have also helped deepen our understanding and characterization of bias in data from a critical stance (e.g., there is never just one bias, bias detection is contextual, bias detection can depend on data modality, biases cannot be eradicated \cite{Alvarez2024PolicyAA}) and to identify challenges not only in modelling bias, but also in relation to the underlying documentation process, primarily on how it should not be fully automatized. 
In developing a tool like our ontology, it is important to aim for a careful balance between an effective, useful and comprehensive vocabulary that supports streamlining documentation tasks, while at the same time, avoid dissuading practitioners from critical thinking when engaging in both documentation and bias analysis. The aim of both of these practices is to mitigate negative consequences arising from the deployment of ML systems. However, it is always possible that unintentionally through enforcing standardization or automation on practitioners, new gateways are created that worsen the problem. Some influencing factors are the lack of experience, domain knowledge, or the right incentives \cite{10.1145/3600211.3604674,10.1145/3442188.3445880,10.1145/3555760}. 
Ultimately, this rapport informs our design choices across all iterations of ontology engineering, makes us aware of the limitations of our technical tool, and creates opportunities for refinement in later versions. 

On the other hand, the scope of our ontology is also informed by the growing body of literature on our topic of interest. In this case, we particularly rely on official reports, as is the NIST Special Publication 1270 \cite{933006}, and by periodically identifying relevant work in order to gather background information for a rich vocabulary of biases- (e.g., \cite{10.1016/j.cviu.2022.103552,10.1145/3588433,10.1145/3457607,Mehrabi2019ASO,olteanu2019social}), while also consider emerging work on this topic published at venues such as: ACM FAccT\footnote{\url{https://facctconference.org/}}, AAAI/ACM AIES\footnote{\url{https://www.aies-conference.com/}}, ACM EAAMO\footnote{\url{https://eaamo.org/}}. Concisely, we pay attention to discerning bias, and its detection measures from fairness notions and their measures, combining keywords such as “machine learning”, “artificial intelligence”, "datasets", "bias", "metrics", and “bias mitigation”. 

\subsection{Doc-BiasO Design}

\begin{figure}[!t]
 \vspace{-0.5cm}
    \centering
{\includegraphics[width=12cm,height=8cm,keepaspectratio]{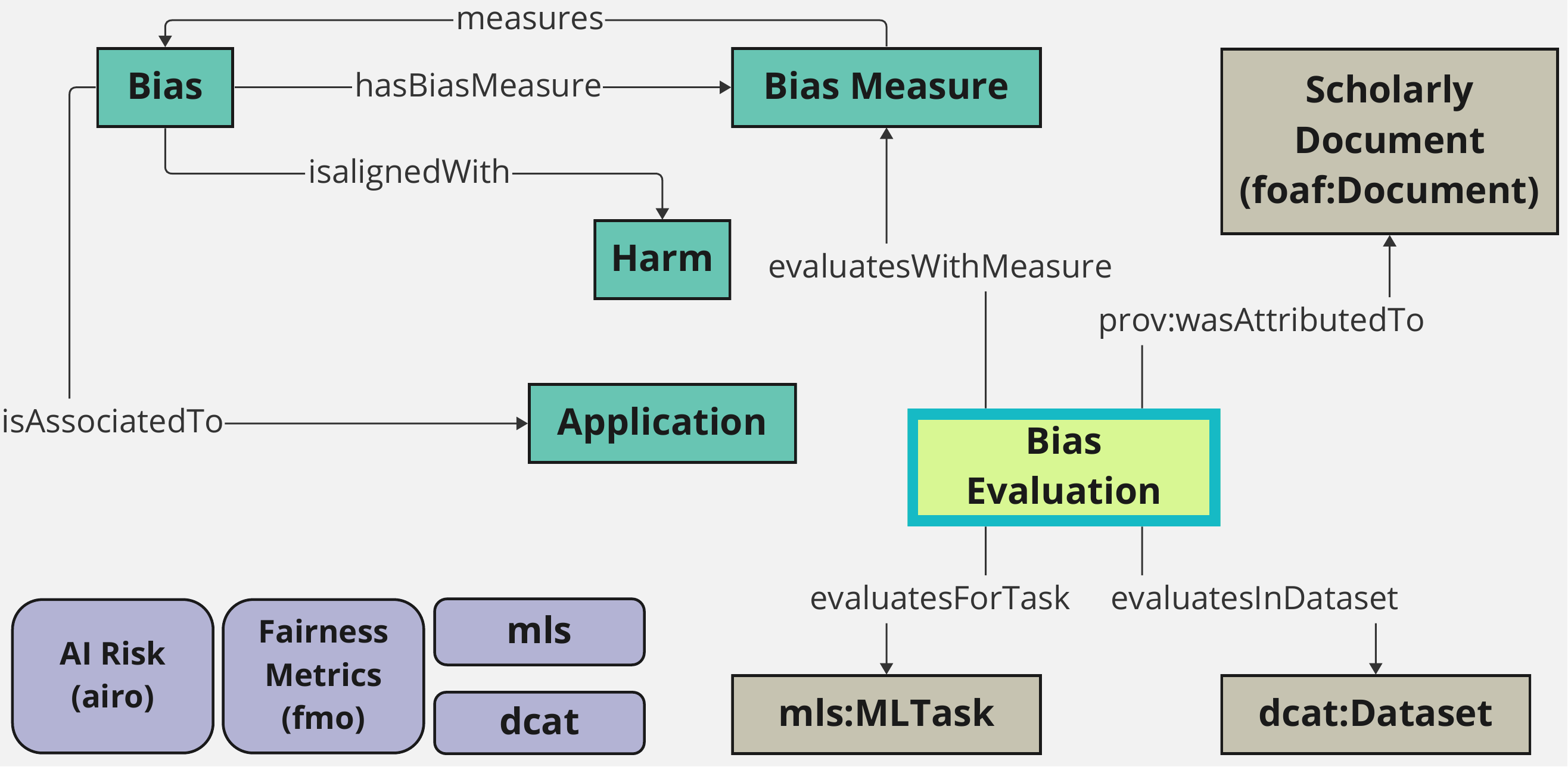}}
    \caption{\textbf{Conceptualization of the Doc-BiasO Ontology.} Core concepts in the ontology are represented as classes, in color-coded boxes, to account for originating vocabularies. While object properties are drawn as directed arrows between classes. In purple colored boxes, relevant and  prominently re-used  vocabularies implemented in the representation of the universe of discourse.}
\vspace{-0.5cm}
\label{fig:doc-bias}
\end{figure}

To design and model our ontology, we adhere to ontology engineering best practices \cite{gruber_toward_1995,DBLP:series/synthesis/2019Kendall}. As such, after the scope is determined and competency questions are defined, re-usable ontologies are identified following a layered approach (i.e., a foundational layer for general metadata and provenance, a domain-dependent layer to cover standards for the relevant area of use, a domain-dependent layer of ontologies specific to our problem of interest) \cite{DBLP:series/synthesis/2019Kendall}.

We first specify the competency questions that emerged during the analysis phase and that represent the intended use of Doc-BiasO: a tool that can be integrated into AI documentation frameworks and that can offer the vocabulary required to characterize these pipelines; ideally, a resource that informs AI practitioners or researchers on the ways in which bias interacts with other components in the AI pipeline, and as a controlled repository as they study the development of a new measure and wish to survey those that already exist.

We then lay the foundation of our ontology by re-using ontologies such as: the SKOS data model \cite{Miles2009SKOSSK}, the PROV data model (PROV-O) \cite{733f89c65e4844f9aabcae1c276a5602}, and the Friend of a Friend (FOAF) vocabulary \cite{Yu2011FOAFFO}. The next layer incorporates standard schemas for data mining and machine learning algorithms, such as the Machine Learning Schema (MLS) ontology \cite{MLSchema}. This schema provides fine-grained descriptions to represent the characteristics and intricacies of ML models. Similarly, the Data Catalog Vocabulary (DCAT) \cite{Albertoni2023TheWD} enables  the fine-grained description of datasets and data services in a catalog using a controlled and rich vocabulary. By extension, the Data Quality Vocabulary (DQV) \cite{Albertoni2020IntroducingTD} provides a framework and vocabulary to assess the  quality of a dataset, offering an extensive catalog of quality metrics. For our third layer, we look at previous work on bias, specifically the BODP \cite{Kaushik2021} and the Artificial Intelligence Ontology (AIO).\footnote{\url{https://bioportal.bioontology.org/ontologies/AIO?p=summary}} The class \texttt{AIO:Bias} is our starting point, which we organize in hierarchies via \texttt{rdfs:SubClassOf}, as per the AIO modelling, and in order to represent different kinds of bias identified in the literature i.e., representation bias, popularity bias, demography bias. We build on the pattern and ontology, however, it does not suffice to our modelling needs. For this reason, all missing concepts are incorporated manually, as we set out to capture and explicitly document otherwise unstated assumptions about bias in relation to ML systems \cite{blodgett-etal-2020-language}. Critical data studies \cite{Shelby2022SociotechnicalHO,blodgett-etal-2020-language} maintain that for bias detection tasks to be meaningful, practitioners must \textit{reflect} on possible harms that can emerge upon the deployment of an ML system in dynamic societal and cultural contexts. Here, we emphasize thus on both, the importance of assisting practitioners via the development of tools that streamline tasks that may be perceived as a burden \cite{10.1145/3442188.3445880}, while avoid dissuading them from reflecting about harms that could emerge from deploying these systems. For that reason, in our modelling we align scoped biases with harms, with the objective to make explicit the articulation of otherwise alleged, unstated negative consequences attributed to ML systems. However, our expectation and recommendation, is that users will enrich the proposed vocabulary with the results derived of their own explorations, in a similar line as with AI incident databases. Furthermore, bias is not singular, and highly context dependent, meaning that most biases are studied and defined in association to a particular ML application. To represent both of these concepts, we model \texttt{bias:Harm} and \texttt{bias:Application}. The central concept in our ontology is \texttt{bias:BiasMeasure}. This class represents a measure defined in some \texttt{foaf:Document}, evaluated in a \texttt{dcat:Dataset} (that has some characteristics), and for a particular \texttt{mls:MLTask}. \texttt{bias:BiasEvaluation} is the class that represents the n-ary relationship between entities schematized in the extended entity relationship model completed at the start of the design phase. Figure \ref{fig:doc-bias} illustrates a conceptual overview of the core classes and relationships of Doc-BiasO. 

\paragraph{Towards a Comprehensive Vocabulary for Trustworthy AI} The Trustworthy AI framework requires a comprehensive formal vocabulary that unifies approaches and contemplates terminology and concepts of ML pipelines, and in broader terms AI, holistically. This type of resource can contribute to the generation of metadata that primes reproducibility and traceability of research results \cite{10.1145/3528574,wilkinson2016fair}, a known issue in ML research and development \cite{10.1145/3531146.3533158,Albertoni2023ReproducibilityOM}. Moreover, it can help achieve a certain degree of standarization for the area. Motivated by this, we perform an analysis of the FMO \cite{10.1145/3514094.3534137} and VAIR \cite{10.1145/3593013.3594050} ontologies, as to determine their characteristics and how they fit into our model. We also do this with the aim to achieve a good balance between ontology re-use and down the line overhead derived from doing so \cite{DBLP:series/synthesis/2019Kendall}. Key takeaways are: \begin{enumerate} \item FMO complements Doc-BiasO by giving coverage to existing fairness metrics used to evaluate ML systems. Specifically, metrics pertaining to machine learning problems of classification and regression; \item VAIR captures a wider scope of AI system deployment to instill accountability on an AI provider (i.e., a party that places the system on the market) and thus capture specifications of risky applications of AI from a regulatory point of view; \item both ontologies represent bias, however, with differing modelling objectives. FMO organizes \texttt{fmo\#Bias} in a hierarchy with seven subclasses, two of these are used in relation to some fairness metric. VAIR represents \texttt{vair\#Bias} as a subclass of \texttt{airo\#Consequence}.\end{enumerate} 
To avoid constraining our modelling, we opt to not import either ontology in its entirety. When needed, we implement OWL axioms to assert class equivalence, i.e., \texttt{owl:equivalentClass}.
Otherwise, we reference external concepts using annotation properties. 

\subsection{Instantiating Doc-BiasO}
To showcase an instantiation of Doc-BiasO, we look at an example based on bias detection in relation to recommender systems, commonly implemented in online social networks.  

\begin{figure}[!t]
\centering
{\includegraphics[width=10cm,height=5cm,keepaspectratio]{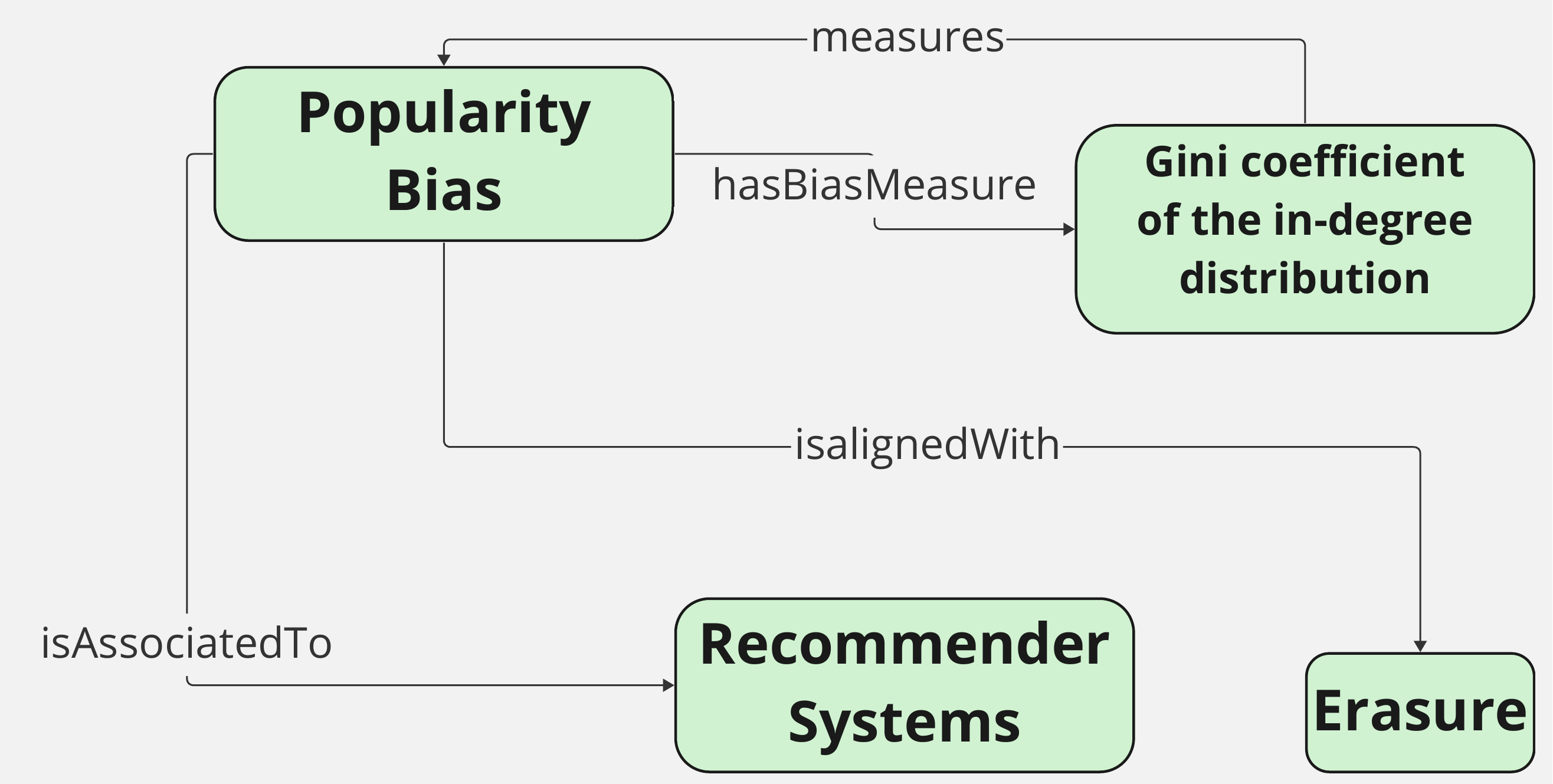}}
\caption{\textbf{Conceptualization of an instance of Doc-BiasO.} Instances of the Doc-BiasO ontology are represented with round-edge boxes and the color green. ``Popularity Bias" is an instance of bias:Bias. Related classes are also exemplified.}
\label{fig:biasinstance}
\end{figure}

The class \texttt{Bias} is instantiated as \texttt{Popularity Bias}. This bias \texttt{is Associated With}, an instance of the class  \texttt{Application}, \texttt{Recommender System} and \texttt{has a Bias Measure}, \texttt{"Gini coefficient of the in-degree distribution"}. In this example, \texttt{Popularity Bias} \texttt{is Aligned With} the instance of the class \texttt{Harm}, which is \texttt{Erasure}. We illustrate this in Figure \ref{fig:biasinstance}.

\section{Evaluation}
\label{sec:eval}

The domain analysis and scope definition of Doc-BiasO, as already described in Section \ref{sec:scope}, derived a set of competency questions that was also used to convey the requirements that would guide the engineering of our ontology. As part of the process, we tested and refined the Doc-BiasO ontology by implementing the formalization of the competency questions originally expressed in natural language as SPARQL queries.\footnote{\url{https://www.w3.org/TR/sparql11-query/}} 
The queries were tested to make sure the results were the expected ones. 

To illustrate their adequacy, we continue with the example introduced earlier, and start by posing Q1 \emph{"Given a particular bias, 
what is its definition?”}; our example uses \texttt{Popularity Bias}. Below the query result: \begin{quote} "When collaborative filtering recommenders emphasize popular items (those with more ratings) over other “long-tail”, less popular ones that may only be popular among small groups of users."@en \end{quote}
\noindent This expected result is expressed as a \texttt{rdfs:Literal} in \texttt{English}. We follow this question by posing Q4.1 \emph{"How many measures have been documented for it?"}. The results produced by executing the corresponding query, specified in Listing \ref{lst:sparql1}, are that for \texttt{Popularity Bias}, we have 3 measures. We choose the measure, \texttt{Gini coefficient of the in-degree distribution}, to learn more about it. We proceed to execute the query that corresponds to Q6. \emph{what is its formalization?}. The corresponding SPARQL query is specified in Listing \ref{lst:sparql2}, with its execution projecting the definition for the chosen measure and the formalization for it in natural language. 

\begin{listing}[!t]
\caption{\textbf{SPARQL Query for Competence Question Q4.1}}
\label{lst:sparql1}
\begin{minted}{sparql}
PREFIX  skos: <http://www.w3.org/2004/02/skos/core#>
PREFIX  owl:  <http://www.w3.org/2002/07/owl#>
PREFIX  rdfs: <http://www.w3.org/2000/01/rdf-schema#>
PREFIX  bias: <https://bias-project.x/bias/>

SELECT DISTINCT  
    ?bias_1 (COUNT(DISTINCT ?biasMeasure_1) AS 
    ?number_of_measures)
WHERE { ?bias_1  rdfs:subClassOf bias:Bias .
        ?biasMeasure_1 bias:measures ?bias_1}
GROUP BY ?bias_1
\end{minted}
\end{listing}

\begin{listing}[!t]
\caption{\textbf{SPARQL Query for Competence Question Q6}}
\label{lst:sparql2}
\begin{minted}{sparql}
PREFIX skos: <http://www.w3.org/2004/02/skos/core#>
PREFIX owl: <http://www.w3.org/2002/07/owl#>
PREFIX rdfs: <http://www.w3.org/2000/01/rdf-schema#>
PREFIX bias: <https://bias-project.x/bias/>

SELECT DISTINCT 
    ?biasMeasure_1 ?definition_1  ?formalization_1 
WHERE { 
   ?biasMeasure_1 rdfs:subClassOf bias:BiasMeasure ;
                       skos:definition ?definition_1 ;
                       bias:formalization ?formal_1
FILTER ( (  REGEX(str(?biasMeasure_1), "Gini", 'i')))}
\end{minted}
\end{listing}

As part of the evaluation process, we also report on the quality of Doc-BiasO. Table \ref{tab:evalresults} summarizes the results obtained according to three indicators defined in \cite{Frber2018LinkedDQ}. 

\begin{table}[!t]
\centering
\label{tab:evalresults}
%\resizebox{.99\textwidth}{!}{%
\begin{tabularx}{.90\textwidth}{lX}
\hline
\textbf{Indicator} & \textbf{Results} \\ 
\hline
{\textbf{Completeness}}\ 
& \\ 
\multirow{1}{*}{Bias} \
&  All 51 subclasses have verifiable definitions based on the NIST report, $ \frac{59}{51}$ =  115\%.  \\
\multirow{1}{*}{Bias Measures} \
&  8 subclasses with verifiable definitions based on ongoing literature review, 24 instances based on 3 case studies.  \\
\hline
{\textbf{Interoperability}}\
& \\ 
{Using external vocabulary} \
&  $ \frac{316}{389}$ = 81\%   \\ 
{Used proprietary vocab}\
&  $ \frac{73}{389}$ = 19\%   \\
{}\
& \\
\hline
{\textbf{Accessibility}}\ 
& \url{http://ontology.tib.eu/DocBIASO/visualization} \\ 
\hline
\end{tabularx}
\caption{\textbf{Quality Indicators for Doc-BiasO.}}
\end{table}

\subsection{Automatic Ontology Evaluation}
This version of Doc-BiasO has also been validated with online tools to verify its consistency and syntactical validity, as well as to check for modelling anomalies or errors. 
First, we checked that our ontology is syntactically correct using the W3C RDF validation service.\footnote{\url{https://www.w3.org/RDF/Validator/}} The results indicated a successful validation of our RDF document. Secondly, we checked for logical consistency by running the DL reasoning engine Pellet (v.2.2.0), as a plug-in for the Protégé open-source platform (v.5.6.1).\footnote{\url{https://protege.stanford.edu/software.php}} We choose this engine as it is a complete reasoner. The results determined that Doc-BiasO is logically coherent and consistent. Finally, we scanned our ontology with the “OOPS! Ontology Pitfall Scanner” \cite{poveda2014oops} to automatically dismiss the existence of modelling pitfalls; the evaluation results were also positive, as there were no bad practices detected by the tool. 

\section{Conclusions and Future Work}
\label{sec:conclusion}
%todo 
In this work, we presented Doc-BiasO, an ontology for bias measures found in the literature that can support the elaboration of documentation of bias in machine learning pipelines. Our objective is to contribute towards improving the interpretation of these pipelines in terms of biases captured, and the derived harms attributed to ML systems. Further, we make a call for a unified controlled vocabulary for the Trustworthy AI framework, and assess existing relevant work. We technically evaluated Doc-BiasO and showcase an example of its instantiation. 
Notwithstanding, our work is not without limitations. Firstly, research on bias in ML, and by extension AI, is a fast-moving field, thus providing adequate and updated coverage with our tool is a challenge. Secondly, bias evaluation are highly complex and context dependent tasks. This means that our modelling cannot account for all potential existing biases, and that in general, bias analysis cannot be fully automated, requiring a human-in-the-loop. Thirdly, our resources are yet to be evaluated by ML practitioners outside a research environment. Nevertheless, the addressed limitations are an opportunity for future work. In particular, we intend to add and expand on aspects left unmodeled in this version, and we will liaise with ML practitioners to evaluate the suitability of our tool in real world scenarios. We will also continue the development of a controlled vocabulary for Trustworthy AI, as this resource can foster effective communication between the different actors involved across the ML pipeline.

\section*{Acknowledgments}
    We thank Guillermo Climent, Sammy Sawischa and Yukti Sharma for their support during this research. Mayra Russo is supported by EU-Horizon 2020 research and innovation programme under the MCSA-grant agreement No. 860630, project: NoBIAS. Maria-Esther Vidal is partially supported by Leibniz Association, program "Leibniz Best Minds: Programme for Women Professors", project TrustKG-Transforming Data in Trustable Insights; Grant P99/2020.

\printbibliography

\end{document}